# Concept-Oriented Deep Learning

Daniel T. Chang (张遵)

*IBM (Retired)* dtchang43@gmail.com

**Abstract:** Concepts are the foundation of human deep learning, understanding, and knowledge integration and transfer. We propose concept-oriented deep learning (CODL) which extends (machine) deep learning with concept representations and conceptual understanding capability. CODL addresses some of the major limitations of deep learning: interpretability, transferability, contextual adaptation, and requirement for lots of labeled training data. We discuss the major aspects of CODL including concept graph, concept representations, concept exemplars, and concept representation learning systems supporting incremental and continual learning.

## 1 Introduction

### 1.1 Human Deep Learning

In human learning, deep learning [1] is an approach that involves the critical analysis of new topics and facts, linking them to already known concepts or forming new concepts, and leads to long term retention of concepts so that they can be used for problem solving in new situations. Deep learning promotes understanding and application for life. This is in contrast to surface learning which is the rote acceptance of facts and memorization as isolated and unlinked facts. It leads to superficial retention of facts and does not promote understanding or long term retention of knowledge.

The major characteristics of deep learning are: aiming for understanding, focusing on concepts, and relating new and previous knowledge. According to the Bloom's taxonomy [2], there are four types of knowledge: factual, conceptual, procedural and metacognitive, and six levels of cognitive processes: remember, understand, apply, analyze, evaluate and create. Factual knowledge, such as topics and facts, is locked in time, place, and/or situation. Factual knowledge does not promote understanding. Concepts [3] (e.g., dog) are general ideas derived or inferred from facts. They are abstract and broad, represented by different instances that share common attributes, universal in application, and timeless. Conceptual knowledge is required for understanding and provides the framework for relating new and previous knowledge.

The ultimate goal of learning is knowledge transfer [2]. Factual knowledge doesn't transfer, but conceptual understanding does. Conceptual understanding is built by abstracting "up" from factual knowledge or examples to understand concepts and the relationships among concepts. Whenever we try to apply our insights from one situation to another we are always abstracting to the conceptual level before our knowledge helps us unlock the new situation. When tasks remain similar to one another, this is known as low-road transfer. To transfer knowledge to dissimilar tasks requires high-road transfer which involves highly generalized concepts. A deep foundation of facts or surface learning is key for deep learning and knowledge transfer. Synergistic thinking, which involves the interaction between the factual and conceptual levels of thinking, is essential for deep learning.

## 1.2 Machine Deep Learning

In machine learning, deep learning has more than one definition. A useful, though narrow, definition [4] is: deep learning is neural networks with a large number of layers and parameters in one of four fundamental network architectures: unsupervised pretrained networks, convolutional neural networks, recurrent neural networks, and recursive neural networks Automatic feature extraction is one of the major facets, and great advantages, that deep learning has.

Deep learning is more broadly defined as feature representation learning in [5, 6]. It uses machine learning to discover not only the mapping from feature representations to output but also the feature representations themselves. To learn features that best represent data, the goal is to separate the factors of variation that explain the data. Deep learning does this by learning successive layers of increasingly meaningful feature representations. The 'deep' in deep learning stands for this idea of deep layers of feature representations. Deep learning is therefore layered feature representation learning. These layered feature representations are generally learned via neural networks.

Deep learning has achieved near-human accuracy levels in various types of classification and prediction tasks including images, text, speech, and video data. However, the current technology of deep learning is largely at the level of surface learning, not deep learning, in human learning, focusing on rote memorization of factual knowledge in the form of feature representations. A deep-learning model [6, 7] is just a chain of simple, continuous geometric transformations mapping one data manifold into another. Anything that needs reasoning is out of reach for deep-learning models. There are other major limitations as well. Firstly, the knowledge learned using deep-learning models cannot be transferred because, as discussed earlier, factual knowledge (feature representations) doesn't transfer, but conceptual understanding does. Secondly, deep-learning models are difficult to understand or interpret [8]. This is to be expected since, as discussed earlier, factual knowledge (feature representations) does not promote understanding. Conceptual knowledge is required for understanding. As a result of the first two limitations, deep-learning models cannot leverage contextual knowledge. They are developed in isolation, within the narrow confine of the specific training data used, and do not support contextual adaptation [7]. Lastly, but not the least, deep-learning models require lots of labeled data for training, which can be hard to come by. This is not surprising since deep learning relies on rote memorization of feature representations to perform classification and prediction tasks. It has no conceptual understanding of the data.

### 1.3 Goal and Outline

From the above discussions it should be apparent that conceptual knowledge learning and conceptual understanding are needed to elevate machine deep learning toward the level of human deep learning. We propose Concept-Oriented Deep Learning (CODL) as a general approach to achieve that goal. CODL is an extension of (machine) deep learning. It extends feature representation learning with concept representation learning and it adds the conceptual understanding capability to deep learning. The major aspects of CODL include: concept graph, concept representations, concept exemplars, and concept representation learning systems. These are discussed in the following sections. The last section provides the summary and conclusion.

## 2 Concept Graph

The Big Book of Concepts [9] states that "Concepts are the glue that holds our mental world together." Without concepts, there would be no mental world. As discussed earlier, concepts [3] (e.g., dog) are general ideas derived or inferred from facts. They are abstract and broad, represented by different instances that share common attributes. A concept may contain a set of attributes that describe the concept and a set of sub-concepts that are components of the concept. Concepts may also be related by relationships. The most common relationships include isA relationships.

Concepts and categories go together [9]. That is, whatever the concept is, there is a category of things that would be described by it. For material things (objects), 'category' is usually referred to as 'class'; for abstract things (entities), 'category' is commonly referred to as 'type'. Thus, concepts denote categories, classes or types, and instances denote things, objects or entities.

Microsoft Concept Graph [10, 11] aims to give machines "common-sense computing capabilities" and an awareness of a human's mental world, which is underpinned by concepts. Microsoft Concept Graph is built upon Probase, which uses the world as its model. The concept graph in Probase is automatically learned from billions of web pages and years' worth of search logs. The core taxonomy of Microsoft Concept Graph contains over 5.4 million concepts. Microsoft Concept Graph also has a large data space (each concept contains a set of instances or sub-concepts), a large attribute space (each concept is described by a set of attributes), and a large relationship space (e.g., "isA", "locatedIn").

The Microsoft Concept Tagging Model [11 – 16], a part of Microsoft Concept Graph, maps text entities into concepts with some probabilities, which may depend on the context texts of the entities. Given an input text entity, it returns a ranked list of concepts. Each concept on the list has a probability denoting the possibility of the text entity belonging to this concept. Besides, some common measures for conceptualization (e.g., Typicality) are provided simultaneously.

CODL uses Microsoft Concept Graph as the common / background conceptual knowledge base and the framework for conceptual understanding, due to its probabilistic nature and extensive scope. Microsoft Concept Graph plays an important role in CODL. Its usage in CODL is discussed in the following sections. However, CODL is not limited to using Microsoft Concept Graph as the common / background conceptual knowledge base. Other comparable system can be used as such.

## 3 Concept Representations

In deep learning, feature representations are generally learned as a blob of ungrouped features. However, an increasing number of visual applications nourish from inferring knowledge from imagery which requires scene understanding. Semantic segmentation is a task that paves the way towards scene understanding. Deep semantic segmentation [17] uses deep learning for semantic segmentation.

Deep semantic segmentation makes dense predictions inferring labels for every pixel. It can be carried out at three different levels:

- Class segmentation: each pixel is labeled with the class of its enclosing object or region
- Instance segmentation: separate labels for different instances of the same class
- Part segmentation: decomposition of already segmented classes into their component sub-classes

CODL extends and generalizes deep semantic segmentation. In CODL, feature representations are always learned semantically segmented in a concept-oriented manner. Concept orientation means that each feature representation is associated with a concept, an instance or an attribute. These concepts, instances and attributes form a concept graph. In addition, the concept graph are generally linked to Microsoft Concept Graph, thus leveraging and integrating with the common conceptual knowledge and conceptual understanding capability provided by Microsoft Concept Graph.

A concept representation consists of a concept, its instances and attributes, and all the feature representations associated with the concept and its instances and attributes. If a concept has sub-concepts, its concept representation also consists of the concept representations of its sub-concepts. Concept representations, therefore, are the same as concept-oriented feature representations, but provide a different view. The latter is data driven and provides a bottom-up view starting from feature representations; the former is concept driven and provides a top-down view starting from concepts. Due to the focus on concepts instead of low-level feature representations, concept representations provide the proper view to work with in CODL.

## 3.1 Supervised Concept Representation Learning

Concept representations can be learned using supervised learning. Similar to deep semantic segmentation [17], discussed above, it can be carried out at different levels:

- Concept level: each feature representation is labeled with the concept that owns the feature
- Instance level: separate labels for different instances of the same concept
- Attribute level: separate labels for different attributes of the same concept
- Component level: decomposition of already learned concept representations into their sub-concept representations

The concept, instance and attribute names used for labeling should be taken from Microsoft Concept Graph, if available. This provides direct link to Microsoft Concept Graph to leverage its common conceptual knowledge and conceptual understanding capability.

## 4 Concept Exemplars

As is the case for deep semantic segmentation, it can be difficult to gather and create labeled concept representation datasets to use for training in supervised learning. Due to the semantically-segmented nature of concepts, a good alternative is to use concept exemplars.

A concept exemplar set is a set of one or more typical instances of a concept, possibly augmented with instances generated from the typical instances using identity-preserving transformations. As an example, for image objects the identity-preserving transformations [18] typically include: scaling, translation, rotation, contrast and color. Each concept is associated with at most one concept exemplar set.

With concept exemplars one can use supervised concept representation learning, as discussed earlier, or unsupervised concept representation learning, as discussed below. Concept exemplars can facilitate incremental and continual learning, to be discussed later.

## 4.1 Unsupervised Concept Representation Learning

Exemplar-CNN [18] is an approach for training a convolutional network using only unlabeled data. It trains the network to discriminate between a set of surrogate classes. Each surrogate class is formed by applying a set of transformations to a randomly sampled 'seed' image patch. The resulting feature representations are not task specific. They are generic and provide robustness to the transformations that have been applied during training. The applied transformations thus define the invariance properties that are to be learned by the network.

Unsupervised concept representation learning uses the same approach as Exemplar-CNN. In CODL, concept exemplars play the role of surrogate classes, and identity-preserving transformations that of applied transformations, in Exemplar-CNN. Whereas surrogate classes are based on randomly sampled 'seed' image patches, concept exemplars are based on semantically distinct, typical instances of concepts. Therefore, we expect unsupervised concept representation learning based on concept exemplars to result in generic, transferable concept representations.

# 5 Concept Representation Learning Systems

Concept representation learning systems provide the platforms and tools for use in CODL. They support supervised concept representation learning as well as unsupervised concept representation learning based on concept exemplars. They also provide access to the common / background conceptual knowledge base, such as Microsoft Concept Graph. The following are major aspects of concept representation learning systems which are important for the success of CODL and its application.

## 5.1 Incremental and Continual Learning

In real-world scenarios, concepts and their associated data are almost always collected in an incremental manner. As such, incremental and continual learning [19] is a critical aspect of CODL. A good concept representation learning system must accommodate new concepts and their associated data that it is exposed to and gradually expands its capacity to predict increasing number of new concepts.

Doing incremental learning using deep neural network faces inherent technical challenges. Neural networks embed feature extraction and classification within the same model. This gives rise to the so-called "catastrophic forgetting / interference" problem [20] which refers to the destruction / modification of existing feature representations learned from earlier data, when the model is exclusively trained with data of new concepts.

Therefore, the challenge is to be able to incrementally and continually learn over time by accommodating new concepts and their data while retaining previously learned concept representations. There are various approaches [19] for incremental and continual learning that mitigate, to different extents, catastrophic interference. The regularization approaches alleviate catastrophic interference by imposing constraints on the update of the neural weights. The dynamic architecture approaches change architectural properties in response to new concepts and their data, either by dynamically accommodating novel neural resources or by re-training with an increased number of neurons or network layers.

A good approach to use in CODL is that of iCaRL [21] (incremental classifier and representation learning), which is a dynamic architecture approach. The approach allows learning in a concept-incremental way: only the training data for a small number of concepts has to be present at the same time and new concepts can be added progressively. Concept-incremental learning has the following properties:

- it is trainable from a data stream in which examples of different concepts appear at intermittent times,
- at any time it provides a competitive multi-concept classifier for all the concepts learned so far, and
- it does not require storing all training data or retraining already-learned concepts whenever new data becomes available.

The main components [21] that enable simultaneous learning of concept representations and concept classifiers in the concept-incremental manner are:

- concept representation learning using knowledge distillation and prototype rehearsal,
- concept exemplar selection / learning based on herding, and
- concept classification by the nearest mean of concept exemplars.

These are discussed below.

Prior to that we note that the deep learning network is used only for concept representation learning and concept exemplar selection, not for classifying new data, which is done using concept classifiers based on concept exemplars. The concept representation learning scheme, using knowledge distillation, addresses the "catastrophic forgetting / interference" problem of incremental and continual learning. The use of concept exemplars, for concept classifiers, avoids the other problem: storing of all training data.

*Concept Representation Learning*

Whenever data for new concepts arrive, the concept representations and concept exemplar sets are updated. Firstly, the network outputs for the existing concepts are stored. Secondly, an augmented training set is constructed, consisting of the (newly available) training examples for the new concepts together with the concepts exemplar sets for the existing concepts. Finally, the deep learning network is trained / updated by minimizing a loss function to output the correct concept indicators for new concepts (classification loss), and for old concepts, to reproduce the scores stored in the first step (distillation loss). The classification loss enables improvements of the concept representations that allow classifying the new concepts well; the distillation loss ensures that the discriminative information learned for existing concepts is not lost while training for the new concepts.

*Concept Exemplars Selection / Learning*

Concept exemplars selection is required for each concept only once, when it is first learned and its training data is available. For each concept, concept exemplars are selected and stored iteratively until the target number is met. In each step of the iteration, one more example of the training set is added to the concept exemplar set, namely the one that causes the average feature vector over all concept exemplars to best approximate the average feature vector over all training examples. Thus, the concept exemplar set is a prioritized list, with concept exemplars earlier in the list being more important. The concept exemplar set may be augmented using identity-preserving transformations, as discussed earlier.

*Concept Classification Based on Concept Exemplars*

For concept classification, a nearest-mean-of-concept-exemplars classification strategy is used. To predict a concept label for a new sample, it first computes a prototype vector for each concept by computing the average feature vector of all concept exemplars for the concept. It then computes the feature vector of the sample that should be classified and assigns the concept label with the most similar prototype. The concept prototypes automatically change whenever the concept representations change, making the classifier robust against changes of the concept representations (as new concepts are learned)

## 5.2 Concept Taxonomy and Basic-Level Concepts

Concept taxonomy [9] is one particular kind of concept organization: the hierarchical structure of concepts with each branch being a sequence of progressively larger concepts in which each concept includes all the previous ones. Different levels of concepts reflect different levels of abstraction, which associate with different attributes. These taxonomic concepts are important for thought and communication.

In a concept taxonomy, the concepts that are higher in the hierarchy are superordinate to the lower-level concepts; the lower-level concepts are subordinate to the higher-level ones. The only relationship allowed between concepts in the hierarchy is the set inclusion relationship: the set of instances of a superordinate concept (e.g., dog) includes the set of instances of its subordinate concept (e.g., bull dog). The set inclusion relationship is called the ''isA'' relationship, which is asymmetric and transitive. The transitivity of concept relationship in the hierarchy leads to a similar transitivity of attribute ascription, called attribute inheritance. Every attribute of a concept is also an attribute of the concept's subordinates.

By being able to locate a concept in its proper place in the concept taxonomy, one can learn a considerable amount about the concept, e.g., its superordinates and inherited attributes. In CODL, this is achieved by accessing Microsoft Concept Graph, as discussed near the beginning. Clearly, this is an important ability, since it allows one to immediately access knowledge (concepts) about new objects or entities without the need to directly learn.

*Basic-Level Concepts*

The objects and entities that we encounter every day do not each fit into a single concept, but can be classified with a large number of different concepts. It is important to know the preferred concept by which people think about any one object or entity.

Any object or entity can be thought of as being in a set of hierarchically organized concepts, i.e., a concept taxonomy, ranging from extremely general (e.g., animal) to extremely specific (e.g., bull dog). Classification at the most general level maximizes accuracy of classification. Most specific concepts, on the other hand, allow for greater accuracy in prediction. Of all the possible concepts in a concept taxonomy to which a concept belongs, a middle level of specificity, the basic level [9], is the most natural, preferred level at which to conceptually carve up the world. The basic level (e.g., dog) can be seen as a compromise between the accuracy of classification at a most general level and the predictive power of a most specific level.

Superordinate concepts (e.g., animal) are distinctive but not informative. Subordinate concepts (e.g., bull dog) are informative but not distinctive. It is only basic-level concepts (e.g., dog) that are both informative and distinctive. Basic-level concept is important because it provides rich information with little cognitive efforts. When a person obtains the basic-level concept of an unfamiliar object or entity, she will associate the object or entity with the known attributes of the basic-level concept.

Therefore, to be effective, in CODL one should focus on learning and using concept representations for basic-level concepts [11]. Superordinate concepts are automatically "learned" by accessing Microsoft Concept Graph, as discussed earlier. When needed, this can be supplemented by learning and using concept representations for selective subordinate concepts.

## 6 Summary and Conclusion

Concepts are the foundation of human deep learning, understanding, and knowledge integration and transfer. The current technology of machine deep learning is largely at the level of surface learning in human learning, focusing on rote memorization of factual knowledge in the form of feature representations. To elevate machine deep learning toward the level of human deep learning, we proposed concept-oriented deep learning (CODL) which extends (machine) deep learning with concept representations and conceptual understanding capability.

CODL leverages Microsoft Concept Graph, or something comparable, as the common / background conceptual knowledge base and the framework for conceptual understanding. In particular, concept names and concept taxonomies (isA relationships) originate from Microsoft Concept Graph. In CODL, feature representations are always learned semantically segmented in a concept-oriented manner. Concept representations are the same as concept-oriented feature representations, but from a top-down, concept-driven perspective which is the focus of CODL. It can be difficult to gather and create labeled concept representation datasets to use for training. Due to the semantically-segmented nature of concepts, a good alternative is to use concept exemplars.

Concept representation learning systems provide the platforms and tools for use in CODL. They support supervised concept representation learning as well as unsupervised concept representation learning based on concept exemplars. Since, in real-world scenarios, concepts and their associated data are almost always collected in an incremental manner, a good concept representation learning system must support incremental and continual learning (using concept exemplars). Also, to be effective, in CODL one should focus on learning and using concept representations for basic-level concepts.

By focusing on learning and using concept representations and concept exemplars, CODL is able to address some of the major limitations of deep learning: interpretability, transferability, contextual adaptation, and requirement for lots of labeled training data.